\documentclass[sigconf]{acmart}

\usepackage{array}
\usepackage{tabularx}
\usepackage{multirow}
\usepackage{xspace}
\usepackage{listings}
\usepackage{makecell}
\usepackage{xcolor}
\usepackage{siunitx}
\newsavebox{\ACMmwbox}
\newcommand{\ACMmaxwidth}[2]{%
  \sbox{\ACMmwbox}{#2}%
  \ifdim\wd\ACMmwbox>#1\relax
    \resizebox{#1}{!}{\usebox{\ACMmwbox}}%
  \else
    \usebox{\ACMmwbox}%
  \fi}

\providecommand{\justifying}{\leftskip=0pt \rightskip=0pt
  \parfillskip=0pt plus 1fil\relax}
\makeatletter
\@ifundefined{FlushLeft}{}{}
\@ifundefined{FlushRight}{}{}
\@ifundefined{Center}{}{}
\@ifundefined{justify}{}{}
\makeatother

\let\ACMorigO\O
\let\ACMorigS\S
\let\ACMorigP\P
\let\ACMorigL\L
\let\ACMorigH\H
\let\ACMorigt\t
\let\ACMorigr\r
\newcommand{\RestoreLaTeXAccents}{%
  \let\O\ACMorigO \let\S\ACMorigS \let\P\ACMorigP \let\L\ACMorigL
  \let\H\ACMorigH \let\t\ACMorigt \let\r\ACMorigr}

\AtBeginDocument{%
  }

\newcommand{\eat}[1]{}

\newcommand{\sstab}{\rule{0pt}{8pt}\\[-2.4ex]}

\newcommand{\bi}{\begin{itemize}}
\newcommand{\ei}{\end{itemize}}
\newenvironment{tbi}{\begin{itemize}
        \setlength{\topsep}{1.5ex}\setlength{\itemsep}{0ex}}
        {\end{itemize}\vspace{-0.5ex}}

\newcommand{\kw}[1]{{\ensuremath{\mathsf{#1}}}\xspace}
 \newcommand{\eop}{\hspace*{\fill}\mbox{$\Box$}}

\newcommand{\be}{\begin{enumerate}}
\newcommand{\ee}{\end{enumerate}}
\newcommand{\beqn}{\begin{eqnarray*}}
\newcommand{\eeqn}{\end{eqnarray*}}

\newcommand{\stitle}[1]{\vspace{1.5ex}\noindent{\bf #1}}

\newcommand{\eetitle}[1]{\vspace{0.8ex}\noindent{\underline{\em #1}}}
\renewcommand{\t}{\tau}

\renewcommand{\r}[1]{{\it rule}(#1)}

\newcommand{\ie}{\emph{i.e.,}\xspace}
\newcommand{\eg}{\emph{e.g.,}\xspace}

\newcommand{\A}{{\mathcal A}}
\newcommand{\Q}{{\mathcal Q}}
\newcommand{\RQ}{{\mathcal{RQ}}}
\renewcommand{\O}{{\mathcal O}}
\renewcommand{\S}{{\mathcal S}}

\renewcommand{\P}{{\mathcal P}}

\newcommand{\M}{{\mathcal M}}

\newcommand{\C}{{\mathcal C}}
\newcommand{\G}{{\mathcal G}}
\renewcommand{\L}{{\mathcal L}}
\renewcommand{\H}{{\mathcal H}}

\newcommand{\comwu}[1]{{\color{red}[com-Wu:~{#1}]}}

\newcommand{\geooutagebench}{\kw{GeoOutageBench}}

\newcounter{example}
\renewcommand{\theexample}{\arabic{example}}
\newenvironment{example}{
        \vspace{1.5ex}
        \refstepcounter{example}
        {\noindent\bf Example \theexample:}}{
        \eop\vspace{1.5ex}}

\newcommand{\nltoq}{\kw{NL2STQ}}
\newcommand{\qtoa}{\kw{Q2A}}
\newcommand{\ontoqtoa}{\kw{OntoQ2A}}
\newcommand{\multiqtoa}{\kw{mmQ2A}}
\newcommand{\multiontoqtoa}{\kw{mmOntoQ2A}}

\copyrightyear{2026}
\acmYear{2026}
\setcopyright{cc}
\setcctype{by}
\acmConference[SIGSPATIAL '26]{The 34th ACM International Conference on Advances in Geographic Information Systems}{November 03--06, 2026}{Riverside, CA, USA}
\acmBooktitle{The 34th ACM International Conference on Advances in Geographic Information Systems (SIGSPATIAL '26), November 03--06, 2026, Riverside, CA, USA}
\acmDOI{10.1145/3841645.3842982}
\acmISBN{979-8-4007-2950-8/2026/11}

\begin{document}
\title
[GeoOutageBench]{GeoOutageBench: Benchmarking Ambiguity-aware, Ontology-grounded Geospatiotemporal KGQA for Multimodal Power Outage and Resilience Analysis}

\author{Ethan D. Frakes}
\orcid{0009-0008-8869-5703}
\authornote{Corresponding author.}
\affiliation{%
  \institution{University of Central Florida}
  \city{Orlando}
  \state{Florida}
  \country{USA}
}
\email{ethan.frakes@ucf.edu}

\author{Amy Kvien}
\orcid{0000-0003-4546-1858}
\affiliation{%
  \institution{University of Central Florida}
  \city{Orlando}
  \state{Florida}
  \country{USA}
}
\email{amy.kvien@ucf.edu}

\author{Rishabh Kundu}
\orcid{0000-0002-0719-3939}
\affiliation{%
  \institution{Case Western Reserve University}
  \city{Cleveland}
  \state{Ohio}
  \country{USA}
}
\email{rxk857@case.edu}

\author{Redad Mehdi}
\orcid{0000-0002-0593-5222}
\affiliation{%
  \institution{Case Western Reserve University}
  \city{Cleveland}
  \state{Ohio}
  \country{USA}
}
\email{mxm1684@case.edu}

\author{Van D. Tran}
\orcid{0009-0008-4355-0543}
\affiliation{%
  \institution{Case Western Reserve University}
  \city{Cleveland}
  \state{Ohio}
  \country{USA}
}
\email{vxt101@case.edu}

\author{Vibha S. Mandayam}
\orcid{0009-0008-8628-9904}
\affiliation{%
  \institution{Case Western Reserve University}
  \city{Cleveland}
  \state{Ohio}
  \country{USA}
}
\email{vsm21@case.edu}

\author{Kristopher O. Davis}
\orcid{0000-0002-5772-6254}
\affiliation{%
  \institution{University of Central Florida}
  \city{Orlando}
  \state{Florida}
  \country{USA}
}
\email{kristopher.davis@ucf.edu}

\author{Erika I. Barcelos}
\orcid{0000-0002-9273-8488}
\affiliation{%
  \institution{Case Western Reserve University}
  \city{Cleveland}
  \state{Ohio}
  \country{USA}}
\email{eib14@case.edu}

\author{Roger H. French}
\orcid{0000-0002-6162-0532}
\affiliation{%
  \institution{Case Western Reserve University}
  \city{Cleveland}
  \state{Ohio}
  \country{USA}}
\email{roger.french@case.edu}

\author{Yinghui Wu}
\orcid{0000-0003-3991-5155}
\affiliation{%
  \institution{Case Western Reserve University}
  \city{Cleveland}
  \state{Ohio}
  \country{USA}
}
\email{yinghui.wu2@case.edu}

\author{Mengjie Li}
\orcid{0000-0002-9531-2474}
\affiliation{%
  \institution{University of Central Florida}
  \city{Orlando}
  \state{Florida}
  \country{USA}
}
\email{mengjie.li@ucf.edu}

\renewcommand{\shortauthors}{Frakes et al.}

\begin{abstract}
\eat{Geospatiotemporal knowledge graphs have become increasingly important for representing mobility, administrative boundaries, raster products, sensor observations, and event dynamics. 
Existing knowledge graph question and answer (KGQA) benchmarks are largely centered on static 
textual facts or general domain reasoning and remain limited in support for 
understanding the usability and capability of LLMs and ontologies in multimodal spatiotemporal KGs, 
particularly for infrastructure outage and resilience analysis. }
We introduce \textbf{GeoOutageBench}, a benchmark for assessing LLM-based geospatiotemporal KGQA for multimodal outage and resilience analysis. Unlike existing KGQA benchmarks for Web knowledge, \geooutagebench considers a spatiotemporal KG that integrates visual, textual, and structured data from outage records, remote sensing, weather observations, storm and power events, geographic entities, and domain ontologies. It provides a competency query taxonomy at different difficulty levels from spatiotemporal containment and proximity, spatiotemporal co-occurrence analysis, multimodal evidence, to hypothetical evaluation. Over multimodal KG and query classes, \geooutagebench provides user-configurable evaluation of three important, highly 
coherent yet less studied  
tasks: (1) LLMs' understanding for ambiguous geospatiotemporal questions in terms of 
NL to SPARQL interpretation, (2) query-driven 
assessment of ontology utility, and (3) 
answer accuracy of multimodal KGQA retrieval.
\geooutagebench provides a design principle and foundation for assessing LLM-KG systems that support real-world infrastructure resilience analysis. Our benchmark, source code, data, results, and other documentation are available at
\url{https://github.com/UCF-SAGE/GeoOutageBench}.
\end{abstract}




\maketitle


\section{Introduction}
\label{sec:introduction}

Question Answering over knowledge graphs (KGQA), which increasingly leverages Large Language Models (LLMs), has significantly advanced knowledge exploration and 
spatial intelligence via natural language (NL) questions. 
\eat{
However, existing KGQA benchmarks remain largely centered on static data, symbolic triples, textual facts, or general-domain reasoning \cite{Aueretal2023,10.1007/978-3-030-30796-7_5, Usbeck20189thCO,Kostenetal2023}. While these generalized datasets have driven progress in semantic parsing, they are insufficient for specialized domains that require complex reasoning over multimodal data, thereby failing to fully stress-test models' spatial and temporal reasoning capabilities.
}
Meanwhile, geospatiotemporal KGs~\cite{Yangetal2024, Bocklingetal2024,Frakesetal2025} have become critical for modeling dynamic spatial systems, including mobility, administrative boundaries, sensor observations, and event dynamics. This calls for effective integration of diverse, dynamic multimodal geospatiotemporal data (\eg localized storm events \cite{NOAAStormEvents}, hurricane tracks \cite{NOAAHurdat}, and nighttime light remote sensing \cite{NASABlackMarble}) to understand evolving geographic phenomena \cite{Yangetal2024, Bocklingetal2024,Frakesetal2025}. 
For example, GeoOutageKG \cite{Frakesetal2025} curates 
factual spatiotemporal data for power outage analysis, remote sensing observations, and energy resilience. 

Effective KGQA requires benchmarking tools to understand how to 
best exploit LLMs, ontologies, and KGs. Existing KGQA benchmarks~\cite{Auer2023SciQA,Kostenetal2023,10.1007/978-3-030-30796-7_5,Usbeck20189thCO}
provide valuable resources for evaluating answer correctness, formal query generation, and execution accuracy over scholarly, encyclopedic, or heterogeneous linked-data KGs. However, they provide limited support for assessing how ontologies, geospatial KGs, and multimodal evidence jointly affect the usability and accuracy of spatiotemporal KGQA for outage and storm-resilience analysis.

\begin{figure}[t]
\centering
\includegraphics[width=0.98\columnwidth]{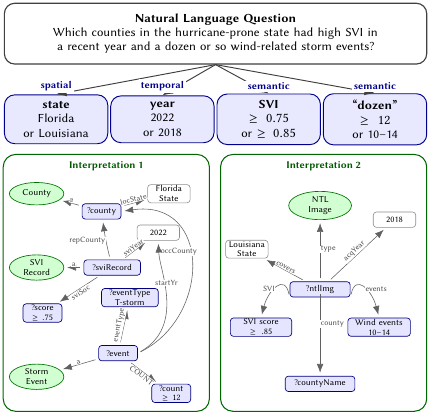}
\Description{A one-column abstract branching diagram for a GeoOutageBench natural-language question. The question points to four underspecified slots labeled as spatial, temporal, semantic, and semantic ambiguity: state, year, SVI threshold, and the meaning of dozen or so storm events. The first expanded interpretation box shows a query graph bound to Florida, year 2022, SVI at least 0.75, Thunderstorm Wind events, and count at least 12. The second interpretation keeps the Louisiana 2018 SVI at least 0.85, Wind-event count between 10 and 14 semantics, but grounds the county path in a nighttime-light image, acquisition year, and VIIRS sensor before joining with SVI and storm-event records.}
\caption{Abstract view of how one question can support multiple KGQA interpretations. The same term can bind the state, year, SVI threshold, and event-count semantics differently, while alternative ontology-compatible traversals can reach the same answer through different 
paths in KG.}
\label{fig:motivation-ambiguity}
\end{figure}

\begin{example}
\label{exa-motivation}
Consider the question
\textit{``Which counties in the hurricane-prone state had high socioeconomic
vulnerability in a recent year and a dozen or so wind-related storm events
recorded that same year?''}
Figure~\ref{fig:motivation-ambiguity} abstracts the interpretations
licensed by this question, with blue elements denoting query variables 
and green elements denoting ontology classes.
For outage and storm analysis, the answer depends on how a KGQA system binds
underspecified state, year, SVI threshold, and event-count semantics. In the
left query reading (Q1), ontology classes ground county, SVI-record, and
storm-event nodes in a direct county-year query; in the right reading (Q2), an
NTL Image class mediates image coverage and acquisition year to the county
answer and associated constraints. These choices
preserve the same county-year co-occurrence structure but lead to different
executable queries and resilience evidence.
\end{example}

Example~\ref{exa-motivation} highlights three limitations of general-purpose KGQA benchmarks that are especially consequential for geospatiotemporal resilience analysis. First, current benchmarks provide limited evidence about whether modern LLMs can perform domain-specific spatiotemporal understanding. Real-world NL questions are often underspecified not only in their entity references, but also in the joint spatial and temporal constraints they imply; as Example~\ref{exa-motivation} shows, a single outage-and-storm question can support multiple valid interpretations over state, year, and event-count semantics. Second, existing evaluations rarely measure the practical \textit{usability} of ontologies and KGs for KGQA: given the same NL workload, how much do a particular ontology and KG improve schema grounding, query generation, and answer accuracy? Third, benchmarks provide limited insight into the role of \textit{multimodality}. Additional modalities such as outage records, storm reports, and remote-sensing imagery may enrich answers but also introduce irrelevant evidence or conflicting signals that degrade KGQA quality.

The need for such efforts is evident to support accurate 
KGQA for location- and temporal-sensitive power event and resilience analysis. Indeed, power outages have become more frequent, more intense, and more prolonged over the past decade, driven by factors including the frequency and severity of extreme weather events, deferred maintenance, and insufficient investment in modernizing energy infrastructure \cite{Li2026}. We advocate a benchmark 
that can provide all the above assessments, beyond ranking LLMs 
over static knowledge, hence helping users to best exploit 
LLMs, ontologies, and multimodal data holistically to support geospatial and temporal retrieval and reasoning.  

\eat{
In this context, ontologies provide the formal vocabulary and relational structure that define how domain concepts, entities, and constraints are represented within a knowledge graph. For our GeoOutageKG knowledge graph, the underlying ontology specifies concepts such as outage records, storm events, satellite observations, counties, temporal intervals, and their spatial or semantic relationships, enabling consistent KG construction and query interpretation. Despite the advancement of these specialized spatial ontologies, there is a distinct lack of tools for benchmarking LLM-based natural language interfaces in these environments. Existing geographic QA datasets are often restricted in scale or do not support the complex topological or temporal interval functions needed for real-world geospatial tasks \cite{Yangetal2023}. Without rigorous benchmarking frameworks, it remains difficult to measure how accurately LLM-KG systems can generate valid spatiotemporal semantic queries or retrieve actionable infrastructure data during rapidly evolving meteorological events. Additionally, while preexisting work has produced knowledge graphs in the energy and outage domains \cite{Koretal2020, Frakesetal2025}, availability of KGQA benchmarks for outage and energy knowledge graphs is limited. 
}

\stitle{Contribution}. 
We introduce \geooutagebench, a domain-specific benchmark for ontology-grounded geospatiotemporal KGQA over multimodal power infrastructure outage data. 
\geooutagebench advances geospatiotemporal KGQA benchmarking for outage and storm resilience through the following primary contributions. 

\sstab 
(1) \textit{A multimodal geospatiotemporal KG}, built on top of the GeoOutageKG~\cite{Frakesetal2025} architecture with multimodal outage records including administrative geographies, storm and hurricane data, social vulnerability indicators, nighttime-light imagery, and outage-severity map metadata.

\sstab 
(2) \textit{A taxonomy of competency queries}, spanning seven categories---spatial containment, spatial proximity, temporal interval, event sequence, spatiotemporal co-occurrence, multimodal evidence, and scenario-based evaluation---with complexity levels ranging from one-hop to multi-hop reasoning. 

\sstab 
(3) \textit{Three benchmark tasks} with evaluation metrics and 
algorithms. \geooutagebench evaluates ontology-grounded NL2SPARQL generation, spatiotemporal KGQA over GeoOutageKG, and query-driven ontology utility. Its metrics capture schema-valid query generation and ambiguity handling, answer quality for multimodal and cross-modal questions, and GeoOutageOnto coverage through workload-vocabulary relevance and schema-utilization density.

\sstab 
(4) \textit{A configurable sandbox platform} that integrates GeoOutageKG and ontologies~\cite{Frakesetal2025}, KG hosting through partners such as GraphDB \cite{GraphDB}, SPARQL engines, and spatial and time-series retrieval tools.


\eat{
    \item A competency query taxonomy spanning seven categories---spatial containment, spatial proximity, temporal interval, event sequence, spatiotemporal co-occurrence, multimodal evidence, and scenario-based evaluation---with complexity levels ranging from one-hop to multi-hop reasoning.
    \item An ontology-constrained NL2SPARQL framework for translating natural-language questions into executable SPARQL \cite{Harrisetal2013} queries through schema linking, operator-aware planning, validation, and repair of ambiguous spatial and temporal expressions.
    \item A query-driven ontology evaluation methodology for measuring whether GeoOutageOnto declares the domain vocabulary required by the SPARQL workload and how densely each query activates that vocabulary \cite{ontocheck}.
    \item A reproducible platform integrating GeoOutageKG \cite{Frakesetal2025}, KG hosting through partners such as GraphDB \cite{GraphDB}, SPARQL querying, and spatial and time-series lookup.
}


\begin{figure*}[t]
\centering
\includegraphics{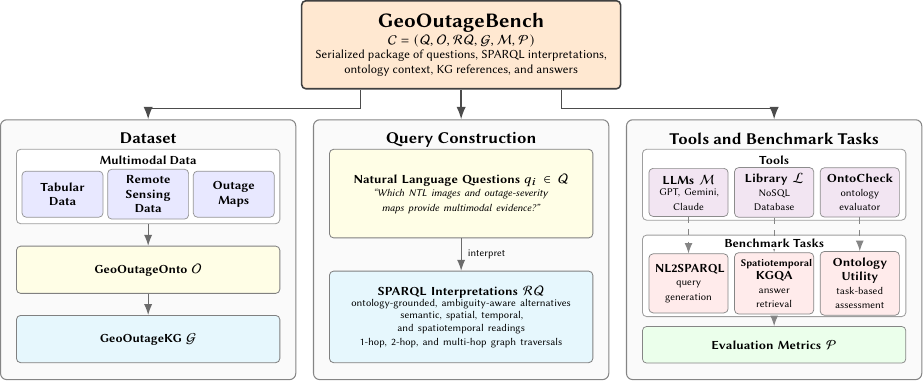}
\Description{Top-down sketch of the GeoOutageBench framework. An orange GeoOutageBench box containing the KGQA profile tuple is centered above three aligned process panels connected by arrows. The Dataset panel contains a Multimodal Data group with text-only boxes for Tabular Data, Remote Sensing Data, and Outage Maps, followed by GeoOutageOnto and GeoOutageKG boxes. The Query Construction panel contains Natural Language Questions and ambiguity-aware SPARQL Interpretations grounded in the ontology, including semantic, spatial, temporal, and spatiotemporal readings with one-hop, two-hop, and multi-hop graph traversals. The Tools and Benchmark Tasks panel contains a Tools group with LLMs, GraphDB, and OntoCheck, a raised Benchmark Tasks group with NL2SPARQL, spatiotemporal KGQA, and ontology-utility assessment, and a custom evaluation metrics box for $\mathcal{P}$ covering ambiguity, schema-linking, spatiotemporal, answer-quality, and ontology-utility metrics.}
\caption{GeoOutageBench framework. Dataset assets are linked through GeoOutageOnto and instantiated in GeoOutageKG; natural-language questions are mapped to ambiguity-aware SPARQL interpretations; and LLMs, GraphDB \cite{GraphDB}, and OntoCheck \cite{ontocheck} support the benchmark tasks and custom metrics $\mathcal{P}$.}
\label{fig:geooutagebench-sketch-upd}
\end{figure*}

\stitle{Related Work}. We summarize related work below. 


\eetitle{Geospatial Ontologies and KGs}.
The W3C Geospatial Ontologies Incubator report \cite{Liebermanetal2007} provides an early foundation for representing geospatial concepts on the Web, including features, feature types, spatial relationships, coordinate reference systems, metadata, and geospatial services. The W3C Basic Geo/WGS84 vocabulary \cite{Brickley2003} offers a lightweight RDF namespace for WGS84 latitude, longitude, and altitude, supporting simple point-based geolocation but not richer geometric or topological reasoning. Related spatial and spatiotemporal ontologies include the Time ontology \cite{w3c_owl_time}, GeoSPARQL \cite{Battleetal2012}, and SOSA/SSN \cite{Janowiczetal2019} for sensor observations, sampled features, and measurement processes relevant to remote sensing and meteorological data. Although widely adopted, these ontologies remain only partially interoperable because their terms are not fully mapped to ISO-standard ontologies and they often separate spatial and temporal domains. The Common Core Ontologies (CCO) \cite{jensenCommonCoreOntologies2024,rudnicki2019overview} address this gap by integrating spatial and temporal concepts in a Basic Formal Ontology-mapped framework \cite{BFO}. As a mid-level ontology, CCO links top-level abstractions to specialized domains, but practical deployment still requires low-level ontologies such as MDS-Onto \cite{Rajamohanetal2025} and GeoOutageOnto \cite{Frakesetal2025}.

\eetitle{KGQA Benchmarks}.
Standard benchmarks such as LC-QuAD 2.0 and QALD-9 focus on open-domain factual questions over encyclopedic knowledge bases such as DBpedia and Wikidata \cite{10.1007/978-3-030-30796-7_5, Usbeck20189thCO}. However, their pattern-based generation may not capture real-world linguistic diversity. Spider4SPARQL \cite{Kostenetal2023} addresses this limitation with a complex benchmark derived from 200 databases, where state-of-the-art models reach only about 45 percent execution accuracy for complex SPARQL queries.

\eetitle{Power System KGQA and Benchmarks}.
Research on KGQA and benchmarking for power outages and resilience analysis remains limited. BuildingQA \cite{Mulayimetal2025}, which provides NL interfaces for building metadata, identifies a significant lexical gap between practitioner terminology and formal graph classes. More broadly, NL ambiguity affects KGQA due to ambiguous schema, vague value mappings, and underspecified language \cite{floratouetal2024}.
Other domain benchmarks include SciQA \cite{Auer2023SciQA} for scholarly knowledge graphs. Unlike benchmarks over simpler encyclopedic sources, SciQA uses the Open Research Knowledge Graph (ORKG) to capture concept drift and varied granularities of scientific literature. While SciQA provides a useful framework for text-rich KGs and some geographic concepts, its geospatial coverage is limited: most of its 2,465 automatically generated questions are restricted to Computer Science.

We remark that our work is not to develop a new KGQA system, but to advocate and demonstrate enhanced assessment capability to enable finer-grained, in-depth experimental analysis; to provide evaluation infrastructure and assessment tools for LLM-based spatiotemporal analysis of power-system resilience.

\vspace{-2ex}
\section{GeoOutageBench Overview}
\label{sec:geooutagebench}

\begin{figure*}[tb!]
    \centering
    \includegraphics[width=0.95\linewidth]{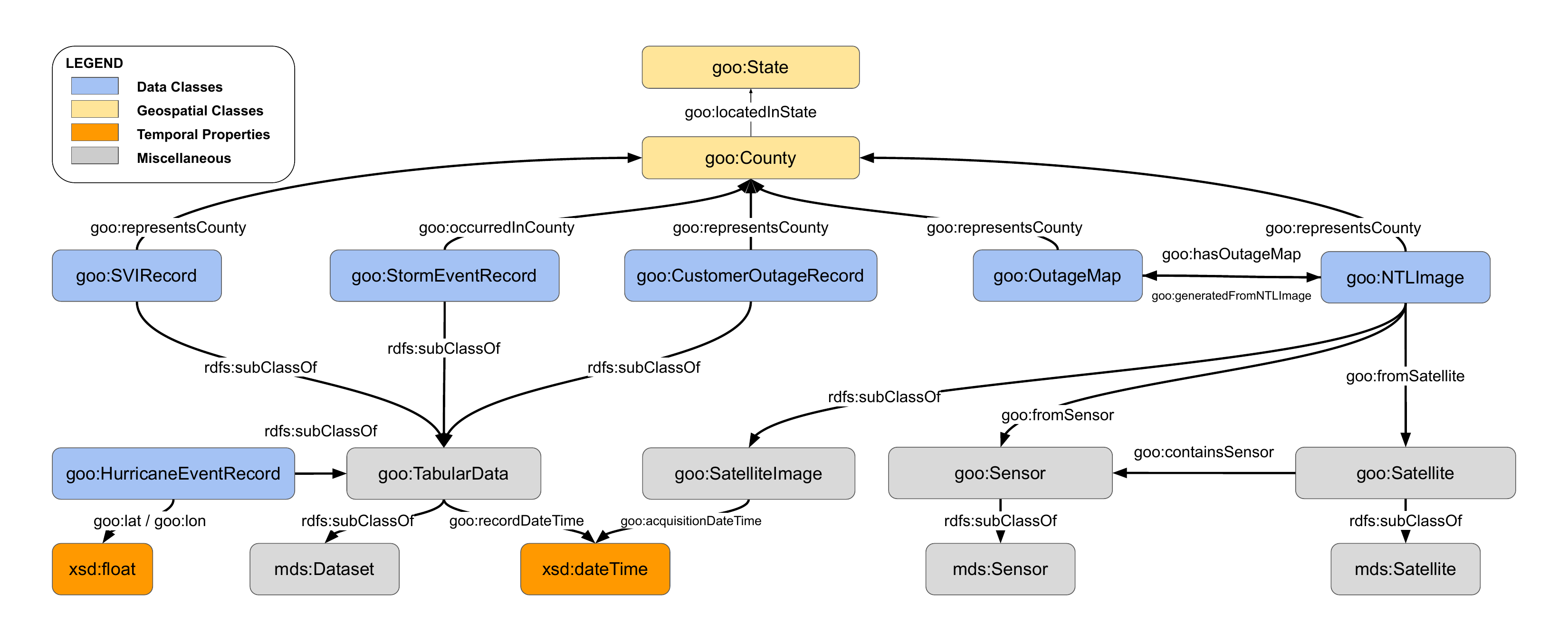}
    \Description{Ontology diagram showing the GeoOutageOnto subset. Data classes are displayed in blue, geospatial classes are in yellow, temporal properties are in orange, and miscellaneous instrument/superclass classes are in gray. \texttt{goo} stands for GeoOutageOnto, \texttt{mds} stands for MDS-Onto, and \texttt{xsd} stands for XML Schema Definition. Edges represent semantic relationships among outage, storm, remote sensing, administrative region, and observation entities.}
    \caption{A fraction of GeoOutageOnto. Data classes are displayed in blue, geospatial classes are in yellow, temporal properties are in orange, and miscellaneous instrument/superclass classes are in gray. \texttt{goo} stands for GeoOutageOnto, \texttt{mds} stands for MDS-Onto \cite{Rajamohanetal2025}, and \texttt{xsd} stands for XML Schema Definition \cite{w3c_xsd_structures_2004}.}
    \label{fig:geooutageonto}
\end{figure*}

\geooutagebench characterizes KGQA assessment using a unified, user-specified \textit{configuration profile}
$\C=(\Q,\O,\RQ,\G,\M,\P)$, where
$\Q=\{q_i\}_{i=1}^{N}$ is a set of NL questions,
$\RQ$ is the corresponding reference SPARQL-interpretation workload,
$\O$ and $\G$ are the ontology and spatiotemporal KG,
$\M$ is a set of language models or agents, and
$\P$ is a set of evaluation metrics.
To support the benchmark tasks (see Section~\ref{sec:benchmark_tasks}), \geooutagebench provides a library $\L$ of primitive operators summarized below, where $'\_'$ denotes an optional wildcard input.
\tbi 
\item \textit{Multimodal Information Extractor}: this operator 
performs data integration and extraction 
to curate and maintain $\G$ from raw multimodal data, 
supported by LLMs, graph data systems, and IE tools. 
\item \textit{NL query interpreter} \nltoq$(q_i,\O,\emptyset,\G,m,\P)$:
takes as input an NL question $q_i$ posed on $\G$, and leverages a language 
model $m\in \M$ to generate structured SPARQL interpretations
$\widehat{\mathcal{RQ}}_{i}^{m}$ that can interpret $q_i$; if $\O$ is specified $(\O\neq\emptyset)$, 
$\O$ is included in prompting $m$ to generate $\widehat{\mathcal{RQ}}_{i}^{m}$ that aligns with the classes,
properties, and constraints in $\O$. 
\item \textit{Graph Query Evaluator} \qtoa$(\_,\_,\RQ,\G,\_,\P)$:
this operator calls a SPARQL engine to deterministically compute the answers of structured interpretations $\RQ(\G)$.
\item \textit{Ontology-based Evaluator} \ontoqtoa$(q_i,\O,\RQ,\G,\_,\P)$:
this operator extends \qtoa by using $\O$ to identify
ontology-compatible labels and relations that satisfy the
conditions expressed in $\RQ$.
\item \textit{Multimodal Evaluators} \multiqtoa (resp. \multiontoqtoa), a variant of 
\qtoa (resp. \ontoqtoa) that supports multimodal retrieval; 
rather than only querying and returning entities in $\G$, 
it performs retrieval over multimodal data and returns a wikipage-style document to include all relevant 
answers. 
\item \textit{User-defined Functions (UDFs) and Tools}:
users register task-specific operators with $\L$ to extend query semantics
beyond what the SPARQL engine evaluates natively. Registered operators include
a geodesic buffer for resolving vague radius expressions, spatiotemporal join,
spatial indexing, kNN search, and raster-tile access for \texttt{NTLImage} and
\texttt{OutageMap} retrieval.
\ei

\vspace{-2ex}
These operators serve as algorithmic building blocks 
and are synthesized for automated 
assessment pipelines in \geooutagebench. 

\vspace{-1ex}
\begin{example}
\label{exa-operator}
For the ambiguous question in Example~\ref{exa-motivation},
\nltoq$(q_i,\O,\emptyset,\G,m,\P)$ produces $\widehat{\RQ}_{i}^{\,m}$, binding the
underspecified state, year, SVI, and event-count phrases to concrete entities and
constraints. \qtoa\ then returns county-level bindings (county, SVI record,
vulnerability score, storm-event count); \ontoqtoa\ checks these against ontology
labels and relations, and \multiqtoa\ and \multiontoqtoa\ attach multimodal evidence to the answers.
\end{example}

\vspace{.5ex}
We next elaborate details of $\O$ and $\G$ (Section~\ref{sec:datasets}), 
competency NL questions $\Q$ and reference SPARQL interpretations $\RQ$ (Section~\ref{sec:query_classification}), 
benchmark tasks with matching evaluation 
metrics $\P$ (Section~\ref{sec:benchmark_tasks}) 
and evaluation methods built on operator library $\L$. Figure~\ref{fig:geooutagebench-sketch-upd} illustrates the general framework. 
While we make a case for power outage and 
storm event analysis, \geooutagebench 
readily extends to other geospatiotemporal 
KGQA query scenarios.

\eat{
This section details the full GeoOutageBench framework, including the datasets and ontology used for knowledge graph construction, the construction of natural language questions, the baseline query set, benchmark tasks, and evaluation metrics. Figure \ref{fig:geooutagebench-sketch-upd} displays a visualization of the GeoOutageBench framework.
}

\subsection{GeoOutageBench Ontologies and KGs}
\label{sec:datasets}

\eat{
Let $\mathcal{O}$ denote the expanded GeoOutageOnto ontology, $\mathcal{G}$ denote GeoOutageKG \cite{Frakesetal2025}, and $\mathcal{D}$ denote the collection of raster, tabular, and administrative-geography datasets instantiated in $\mathcal{G}$. This section describes how each $d \in \mathcal{D}$ is processed, semantically linked through $\mathcal{O}$, and used to support GeoOutageBench questions in multimodal outage detection and storm resilience.
}

\stitle{Multimodal GeoOutageOnto $\O$}. 
\label{sec:geooutageonto}
GeoOutageOnto provides the schema for GeoOutageKG and the \geooutagebench KGQA dataset. The ontology $\O$ is designed as a robust representation for integrating heterogeneous geospatial and temporal resources, including satellite-derived nighttime light images \cite{NASABlackMarble}, meteorological remote-sensing products \cite{NOAAGoes,NASAAqua,NASATerra}, county-level customer outage records \cite{DOEEaglei}, and storm event reports \cite{NOAAStormEvents}. By defining a unified semantic layer over these resources, GeoOutageOnto supports systematic ingestion, alignment, and reasoning for power outage analysis and storm resilience applications.

\begin{example}
\label{exa-geoonto}
Figure~\ref{fig:geooutageonto} illustrates a representative subset of GeoOutageOnto.
Classes associated with ingested datasets are spatially grounded through their administrative divisions, such as counties and states in the U.S. These classes are further organized by modality, distinguishing between quantitative tabular datasets and geospatial raster imagery. GeoOutageOnto is curated to be conceptually well aligned with the Materials Data Science Ontology (MDS-Onto) \cite{Rajamohanetal2025} to promote the standardization of location- and time-sensitive concepts. 
MDS-Onto is aligned with the Basic Formal Ontology (BFO) \cite{BFO}, which justifies the well-groundedness of GeoOutageOnto. 
\end{example}

\stitle{Multimodal Spatiotemporal KG $\G$}. 
\label{sec:geooutagekg}
\geooutagebench specifies $\G$ by default as GeoOutageKG \cite{Frakesetal2025}, enriched with multimodal data sources typed by $\O$. Its semantically enriched entities are grounded by real-world geographic location, acquisition time, and record-time metadata, forming the geospatial, temporal, and multimodal evidence base for \geooutagebench. 
Full statistics on GeoOutageKG classes and instance counts are provided in Table~\ref{tab:kg-stats}.

\begin{table}[H]
    \caption{GeoOutageKG instance and triple counts by class.}
    \label{tab:kg-stats}
    \centering
    \sisetup{group-separator={\,}, group-minimum-digits=4}
    \begin{tabular}{l S[table-format=8.0] S[table-format=9.0]}
        \toprule
        \textbf{Class} & {\textbf{Instances}} & {\textbf{Triples}} \\
        \midrule
        \textit{CustomerOutageRecord} & 11733064 & 117330640 \\
        \textit{NTLImage}             &   335469 &   7907444 \\
        \textit{OutageMap}            &   191661 &   2108267 \\
        \textit{StormEventRecord}     &  2012856 &  36211668 \\
        \textit{HurricaneEventRecord} &     3266 &    128831 \\
        \textit{SVIRecord}            &    21997 &    395940 \\
        \midrule
        \textbf{Total}                & 14298313 & 164082790 \\
        \bottomrule
    \end{tabular}
\end{table}

\eetitle{Raster layer}. 
The raster layer uses NASA Black Marble nighttime light (NTL) observations \cite{NASABlackMarble}, which provide nightly imagery from 2012 to 2025 at 500 meters per pixel; because artificial illumination is a spatially explicit proxy for electricity availability, infrastructure functionality, and human activity, departures from historical radiance can indicate outage impacts when interpreted with administrative boundaries, outage records, and storm metadata.

Raw rasters are partitioned into U.S. counties using county-level bounding boxes and coordinate polygons, then transformed into outage severity maps by comparing daily radiance against the preceding three-month average \cite{Frakesetal2025,Aparcedoetal2024,Coleetal2017,Cuietal2023,Kalbetal2018}. These maps encode anomalous radiance loss as outage severity, allowing $\G$ to represent the spatial extent, intensity, and temporal evolution of disruption and recovery; daily NTL images and outage severity maps are serialized as \texttt{NTLImage} and \texttt{OutageMap} instances, while the observing Suomi-NPP satellite and VIIRS sensor are represented as instances of \texttt{Satellite} and \texttt{Sensor}, 
respectively. 

\begin{figure}[t]
\centering
\includegraphics[width=0.98\columnwidth,totalheight=0.78\textheight,keepaspectratio]{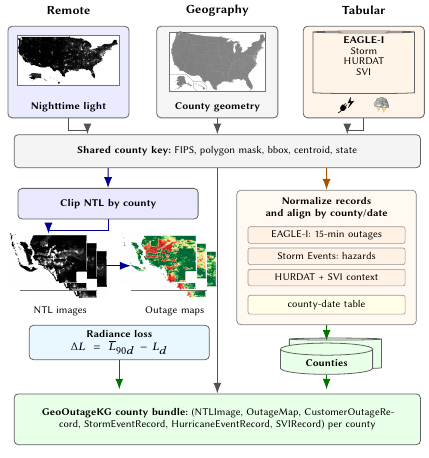}
\Description{One-column multimodal data workflow for GeoOutageBench. Remote-sensing, geographic, and tabular sources feed a shared county key. The raster branch clips nighttime-light imagery by county, places the NTL image stack and outage-map stack side by side with a rightward arrow between them, and shows radiance loss below the stacks. The tabular branch uses one combined normalization and county-date alignment block for outage, storm, hurricane, and vulnerability records, then passes through overlapping cylindrical Counties boxes. Both branches instantiate county-level GeoOutageKG evidence bundles.}
\caption{Multimodal data construction for GeoOutageBench. County geography provides the shared spatial key used to segment nighttime-light imagery, derive radiance-loss outage maps, and ground outage, storm, hurricane, and vulnerability records as county-level GeoOutageKG evidence.}
\label{fig:geooutagebench-data}
\end{figure}

\eetitle{Tabular Layer}. 
The tabular layer comprises quantitative datasets aligned by observation time and grounded through coordinate pairs or administrative divisions, most commonly U.S. counties; currently, the multimodal geospatiotemporal KG $\G$ integrates:

\sstab 
(1) The Department of Energy's EAGLE-I Customer Outage Record \cite{DOEEaglei,DOEEaglei2023,DOEEaglei2024,DOEEaglei2025,brelsford2024dataset}, which provides county-level outage observations at 15-minute intervals (2014-2025) and the primary temporal signal for modeling outage severity and persistence.

\sstab 
(2) NOAA's Storm Events Database \cite{NOAAStormEvents}, which records county-level storm events and damaging episodes from 1950 to 2025, contextualizing outages through meteorological events, affected locations, severity, and infrastructure disruption.

\sstab 
(3) NOAA's Hurricane Databases (HURDAT) \cite{NOAAHurdat}, which contain metadata for tropical cyclones in the North Atlantic Ocean from 1851 to 2025 and the Northeastern Pacific Ocean from 1949 to 2025, including name, maximum wind strength, barometric pressure, and six-hour status updates with location and wind strength.

\sstab
(4) The CDC/ATSDR Social Vulnerability Index (SVI) \cite{CDCSVI}, which provides spatially resolved measures of relative social vulnerability across socioeconomic status, household characteristics, racial and ethnic minority status, etc.; its theme-specific and overall percentile rankings allow GeoOutageKG to incorporate community vulnerability when analyzing outage exposure, disaster preparedness, and post-event recovery capacity. 

\eetitle{Geographic Metadata}. 
$\G$ also serializes metadata for all 3,144 U.S. counties and county-equivalent entities, together with all U.S. states and territories, using the U.S. Census Bureau's TIGER digital mapping and shapefile database \cite{CensusTIGER}. County and state instances include centroid coordinates, Federal Information Processing Standards (FIPS) codes, geohashes, and bounding-box geometries. 
We note that Storm Events, SVI, HURDAT, and TIGER geometries are ingested for all counties, while Black Marble and EAGLE-I derived assets are currently sampled within Florida due to raster volume. Because all layers are partitioned by the same shared county key, extending coverage requires re-running the ingestion pipeline over additional TIGER geometries rather than revising GeoOutageOnto or the query templates.


\newcommand{\amb}[2]{\textbf{#1}\,[\textit{#2}]}

\begin{table*}[htbp]
\centering
\caption{Query classification by primary category. Coverage entries are
\textit{template instances\,/\,SPARQL interpretations} and are inclusive:
an interpretation at a given hop depth also satisfies shallower depths.
Ambiguous spans in the example questions are shown as
\amb{surface form}{binding type}.}
\label{tab:query-classification}
\scriptsize
\setlength{\tabcolsep}{3pt}
\renewcommand{\arraystretch}{1.15}
\begin{tabularx}{\textwidth}{@{}
>{\raggedright\arraybackslash}p{1.55cm}
>{\centering\arraybackslash}p{0.95cm}
>{\centering\arraybackslash}p{0.95cm}
>{\centering\arraybackslash}p{0.95cm}
>{\centering\arraybackslash}p{1.05cm}
>{\raggedright\arraybackslash}X
>{\raggedright\arraybackslash}p{2.15cm}@{}}
\toprule
& \multicolumn{3}{c}{\textbf{Coverage by hop depth}} & & & \\
\cmidrule(lr){2-4}
\textbf{Category} & \textbf{1-hop} & \textbf{2-hop} & \textbf{Multi-hop}
& \makecell[c]{\textbf{Total}\\\textbf{Q\,/\,I}}
& \textbf{Example question} & \textbf{Underlying operation} \\
\midrule

\textbf{Spatial containment}
& 12\,/\,13 & 7\,/\,8 & 7\,/\,8 & 12\,/\,13
& Which \amb{U.S. county}{jurisdiction type} is depicted in the
\amb{power outage severity map}{map type} corresponding to
\amb{a specific customer outage record}{outage record}?
& Region containment, jurisdiction lookup, or spatial aggregation \\
\addlinespace[2pt]

\textbf{Spatial proximity}
& 4\,/\,9 & 4\,/\,9 & 4\,/\,9 & 4\,/\,9
& On \amb{Ian's Florida landfall day}{event date}, what was the peak outage
count in \amb{the southwest Florida county centered on Fort Myers}{place
descriptor}, and were any hurricanes active within \amb{the general area of
the county}{distance/radius abstraction} that day?
& Buffer, distance filter, spatial join \\
\addlinespace[2pt]

\textbf{Temporal interval}
& 11\,/\,14 & 3\,/\,3 & 2\,/\,2 & 11\,/\,14
& How many and which \amb{hurricanes}{cyclone type} made landfall in
\amb{Louisiana}{jurisdiction} in \amb{August}{month} across all years?
& Temporal filtering, aggregation, and interval comparison \\
\addlinespace[2pt]

\textbf{Event sequence}
& 3\,/\,7 & 3\,/\,7 & 3\,/\,7 & 3\,/\,7
& Which \amb{storm events}{event type} in \amb{Florida}{state} started before
same-county \amb{EAGLE-I records}{outage record source} that later exceeded
\amb{50,000 outages}{outage threshold} on \amb{the same day}{temporal
alignment}?
& Temporal precedence over event graph \\
\addlinespace[2pt]

\textbf{Spatiotemporal co-occurrence}
& 3\,/\,7 & 3\,/\,7 & 2\,/\,4 & 3\,/\,7
& Which \amb{counties}{jurisdiction type} in \amb{the hurricane-prone
state}{state descriptor} had \amb{high socioeconomic
vulnerability}{vulnerability descriptor} in \amb{a recent year}{relative year
descriptor} and \amb{a dozen or so wind-related storm events}{event-count
condition} recorded \amb{that same year}{temporal alignment}?
& Spatial--temporal join \\
\addlinespace[2pt]

\textbf{Multimodal evidence}
& 13\,/\,18 & 6\,/\,10 & 2\,/\,4 & 13\,/\,18
& Which \amb{outage map}{map type} corresponds to \amb{this customer outage
record}{outage record}?
& KG retrieval plus file, text, or image lookup \\
\addlinespace[2pt]

\textbf{Scenario / what-if}
& 2\,/\,4 & 2\,/\,4 & 2\,/\,4 & 2\,/\,4
& If \amb{Fort Myers's county}{focal place} outage impact extended to
\amb{neighboring counties}{spatial relation} during \amb{Hurricane Ian}{event
window}, which neighboring counties had \amb{high social
vulnerability}{vulnerability condition} and \amb{high recorded
outages}{outage condition}?
& Constrained scenario retrieval and impact exploration \\

\midrule
\textbf{Overall}
& \textbf{48\,/\,72} & \textbf{28\,/\,48} & \textbf{22\,/\,38}
& \textbf{48\,/\,72}
& \multicolumn{2}{@{}>{\raggedright\arraybackslash}p{9.4cm}@{}}{%
48 benchmark questions drawn from 11 template families, with 72 reference
SPARQL interpretations. Per-category counts are illustrative; corpus-level
metrics are the reportable figures.} \\
\bottomrule
\end{tabularx}
\end{table*}

\stitle{Multimodal Data Integration}. 
Figure~\ref{fig:geooutagebench-data} summarizes the multimodal data-processing workflow (conducted by multimodal information extractor; Section~\ref{sec:geooutagebench}): geographic boundary data partitions remote-sensing imagery into county-level NTL images and derived outage severity maps while also spatially grounding tabular records as geospatially segmented county-level data. By linking county-partitioned raster imagery and temporally aligned tabular records through shared administrative geographies, $\G$ provides the semantic substrate for GeoOutageBench questions that require multimodal and geospatiotemporal reasoning over power outages, environmental hazards, and community resilience. 

\subsection{Competency Query Classes}
\label{sec:query_classification}

\geooutagebench considers a set of built-in 
competency NL questions $\Q$ for evaluation tasks. 
Questions in $\Q$ are constructed through human authoring and LLM-assisted generation. 

\stitle{NL query classes}. \geooutagebench 
characterizes the query complexity of 
NL questions in terms of the common 
practice of ``hops'' as in Web KGQA 
(length of traversal across entities via 
relations in KG), and spatiotemporal semantics. 
(1) One-hop queries involve direct retrieval or filtering over a single class or relation, while two-hop and multi-hop queries require the composition of multiple relations across linked entities. 
(2) Each NL $q\in \Q$ is further assigned to one of seven categories. Together, these categories address a useful fragment of reasoning needed for outage and storm resilience analysis, from basic spatial or temporal filtering to complex linking of administrative geographies, hazard events, outage records, remote-sensing imagery, and vulnerability indicators.

\stitle{NL query generation}. The built-in 
competency question set $\Q$ contains 
manually designed questions and 
LLM-generated ones. It 
starts with a set of parameterized NL templates 
that specify hop and query category, with 
``placeholders'' for specific targeted 
entities and predicates. 
(1) The human expert team consists of 11 researchers, with overlapping expertise spanning 4 environmental and energy system experts,
4 computer scientists, 2 geospatial experts, and 3 power domain experts. \footnote{Some researchers occupy multiple positions.}
The geospatial and power experts are requested to 
propose practical NL questions, and the full team works together to make ground-truth 
interpretations. 
(2) For LLM-assisted question generation, an initial ``draft'' is generated from GeoOutageBench metadata and ontology conforming to the target query categories and template, and then manually
augmented to introduce varied ambiguity and query-generation difficulty by human experts.

\eetitle{NL Query Interpretation}. For the generated competency NL questions, 
the CS experts provide validated ground-truth (``gold'') SPARQL interpretation sets $\RQ_i$ for each NL question $q_i\in\Q$. The resulting reference query workload is $\RQ=\bigcup_{i=1}^{N}\RQ_i$, and the ground-truth answers $\A=\RQ(\G)$ are derived by executing these queries on supporting multimodal and NoSQL data engines such as graph and document data systems (\eg GraphDB, MongoDB), with their answers passing the manual validation of domain experts. 

Table~\ref{tab:query-classification} summarizes the query categories, hop-complexity, question and interpretation counts, representative questions, and underlying operations used to process the query workload.

\begin{table*}[t]
\centering
\caption{Major benchmark tasks and Evaluation metrics}
\label{tab:evaluation-metrics}
\scriptsize
\setlength{\tabcolsep}{3pt}
\renewcommand{\arraystretch}{1.08}
\ACMmaxwidth{\textwidth}{%
\begin{tabularx}{\textwidth}{p{2.25cm}p{3.10cm}p{5.15cm}X}
\toprule
\textbf{Benchmark Tasks} & \textbf{Evaluation object} & \textbf{Example Metrics in $\P$} & \textbf{What the metrics measure} \\
\midrule
\textbf{Task 1: Geospatiotemporal NL to SPARQL}
& Predicted interpretations $\widehat{\RQ}_{i}^{\,m}$ compared with reference interpretations $\RQ_i$
& Syntax; Compat.; Exact; Alignment; Class F1; Prop. F1; Schema F1; Spatial F1; Temporal F1; ST F1
& Whether $m$ generates executable, ontology-grounded SPARQL and preserves schema, spatial, and temporal constraints under ambiguity. \\
\textbf{Task 2: Ontology Utility}
& Workload term set $T_a$ compared with ontology term set $T_o$
& Relevance / Query Coverage $\frac{|T_a \cap T_o|}{|T_a|}$; Accuracy / Query Closeness $\frac{|T_a \cap T_o|}{|T_o|}$
& Whether $\O$ declares the GeoOutageOnto vocabulary required by the SPARQL workload and how densely each query activates that vocabulary. \\
\textbf{Task 3: Multimodal Spatiotemporal KGQA}
& Predicted answer $\widehat{\A}_{i}^{\,m}$ vs. ground-truth answer $\A_i$ over $\G$
& Prec.; Rec.; Answer F1; Sel.; Hits@$k$; MRR; nDCG@$10$; spatial correctness $S_i$; temporal correctness $T_i$; SRS
& Whether the executable query returns correct entities, values, records, images, or aggregates and preserves the question's spatiotemporal intent. \\
\bottomrule
\end{tabularx}%
}%
\end{table*}

\section{Benchmark Tasks and Evaluation Metrics}
\label{sec:benchmark_tasks}

Given the benchmark-level configuration 
$(\Q, \O, \RQ, \G, \M, \P)$, 
an ML oracle (\eg an LLM) $m\in \M$ 
deterministically maps each NL question $q_i\in \Q$ to a generated set of \textit{interpretations} (\ie syntactically validated structured SPARQL queries) $\widehat{\RQ}_{i}^{\,m}$. 
The corresponding generated query workload over $\Q$ is $\widehat{\RQ}^{\,m}=\bigcup_{i=1}^{N}\widehat{\RQ}_{i}^{\,m}$. 
Let the ground-truth interpretations (resp. answers) of 
$q_i\in \Q$ be $\RQ_i$ (resp. $\A_i=\RQ_i(\G)$, \ie
the union of the answers of each ground-truth SPARQL query 
in $\RQ_i$). The 
ground-truth queries for $\Q$ 
and answers, denoted as 
$\RQ=\bigcup_{i=1}^{N}\RQ_i$ and $\A=\RQ(\G)=\bigcup_{i=1}^{N}\A_i$, 
are defined similarly.
Table~\ref{tab:evaluation-metrics} summarizes the metrics used by the three benchmark tasks 
(see formal 
definitions in Appendix~\ref{app:evaluation-metrics}.) 


\vspace{-2ex}
\subsection{Task 1: Assessing LLM's ability in Geospatiotemporal Understanding} 
\label{sec:task1}

The first benchmark task assesses LLMs $\M$'s 
ability of 
geospatiotemporal understanding, by quantifying the quality of its query interpretation. 
We introduce three groups of 
metrics below, in terms of 
interpretation ambiguity,
query accuracy, and 
accuracy of answers 
from interpreted queries 
compared with ground-truth 
SPARQL queries and answers. 

\stitle{Interpretation Ambiguity}. 
For each question $q_i\in\Q$, the evaluator records the number of ground-truth interpretations $|\RQ_i|$ and generated interpretations $|\widehat{\RQ}_{i}^{\,m}|$; questions with multiple interpretations indicate ambiguity in the NL-to-SPARQL mapping.

\stitle{Query Accuracy}. We measure the accuracy of model $m$ by comparing generated SPARQL interpretations $\widehat{\RQ}_{i}^{\,m}$ with ground-truth interpretations $\RQ_i$. The evaluator reports normalized exact original-query match, interpretation alignment, syntactic validity, and endpoint compatibility with $\G$. In the Task 1 workflow, interpretation alignment is the structural pair-selection score used for each candidate SPARQL interpretation generated by \nltoq per $q_i$. For each generated--gold pair $(\hat{r},r)\in\widehat{\RQ}_{i}^{\,m}\times\RQ_i$, the evaluator extracts class, property, spatial-constraint, and temporal-constraint feature sets, computes the corresponding F1 scores, and takes the maximum average score over all pairs (interpretations) for $q_i$. Thus, alignment measures whether an interpretation preserves the ontology grounding and spatiotemporal constraints of a gold counterpart, rather than whether its syntax exactly matches a gold counterpart.

\eetitle{Schema-level F1 measures}. 
\geooutagebench also provides schema-level 
query accuracy by extracting semantic 
components from 
interpreted and ground-truth SPARQL queries, and reporting F1 scores for classes, properties, schema-linking, spatial constraints, temporal constraints, and spatiotemporal constraints. For ambiguous questions, every model-generated interpretation is scored against its ground-truth counterpart, and the best aligned generated--gold pair provides the schema-level scores. 
This group includes three metrics: class F1, property F1, and schema-linking F1. Class and property F1 compare the classes and object or datatype properties in ground-truth and interpreted queries. Schema-linking F1 is the harmonic mean of class and property F1, penalizing outputs that recover only entities or relations. 

\stitle{Quality of Query Answers}. \geooutagebench provides a 
rich set of evaluation 
metrics for quality of query answers, including (1) commonly used precision, recall, and F1 measures, 
and (2) Hits@$1$, Hits@$5$, Hits@$10$, MRR, nDCG@$10$, spatial correctness, temporal correctness, and SRS, to distinguish 
the assessment of entity-level correctness and spatiotemporal-focused interpretation. 

\geooutagebench can also be easily configured to perform in-depth, finer-grained  
assessment 
for spatiotemporal reasoning, with performance metric $\P$ 
specified as follows. 

\eetitle{Spatiotemporal-focused Metrics}. For spatial and temporal reasoning, $\P$ 
specifies three F1 scores for individual 
competency query classes. This 
includes spatial-constraint, temporal-constraint, and spatiotemporal-constraint. Spatial-constraint F1 captures 
constraints such as geometry vectors, coordinates, distance functions, and spatial predicates; temporal-constraint F1 captures temporal literals, datatype use, date operators, and interval filters. 
Spatiotemporal-constraint F1 is the harmonic mean of spatial and temporal F1. 

\eat{
For each selected pair, spatiotemporal-constraint F1 is:
\begin{equation}
\mathrm{STF1}_q = 
\begin{cases}
\frac{2 \cdot SF1_q \cdot TF1_q}{SF1_q + TF1_q}, & \text{if } SF1_q + TF1_q \neq 0, \\
0, & \text{otherwise}
\end{cases}
\end{equation}
}

The benchmark reports the macro-average of $STF1_i$ across questions and the corpus-level harmonic mean of macro spatial- and temporal-constraint F1. The question-level score is stricter because it requires both constraint types to be present within the same selected interpretation, rather than aggregating them independently across the benchmark.
We provide a detailed description of 
the metrics in Appendix~\ref{app:evaluation-metrics}.

\stitle{Evaluation method}. The assessment is 
triggered by a configuration $(\Q, \O, \emptyset, \G, \M, \P)$, with $\P$ specified as ambiguity and 
accuracy (a vector of accuracy metrics), by default.  \geooutagebench invokes operator \nltoq$(q_i,\O,\emptyset,\G,m,\P)$ (see Appendix~\ref{app:nl2sparql-prompt})
with $q_i$ ranging over $\Q$ and 
$m$ over $\M$ to produce an assessment 
report.

\subsection{Task 2: Ontology Utility Evaluation}
\label{sec:task2}
Task 2 aims to evaluate whether an ontology $\O$ (e.g.,
GeoOutageOnto) is usable and useful by measuring how well it supplies
the domain vocabulary required by the reference interpretations
$\RQ$ or model-generated interpretations
$\widehat{\mathcal{RQ}}^{m}$.

\stitle{Evaluation Metric}. Given an interpretation workload
$\RQ$ (or $\widehat{\mathcal{RQ}}^{m}$), let $T_a$ denote the
ontology terms occurring in the supplied SPARQL interpretations.
For a scope radius $h$, we construct the scoped ontology term set
$T_o$ as follows: for each class label $l\in T_a$ with an instance
$v\in\G$ typed by $l$, $T_o$ includes all ontology terms in $\O$
associated with nodes reachable within $h$ hops of $v$.
Intuitively, $T_o$ includes 
all the concept classes that may contribute in 
answering $\Q$ within a scope 
specified by $h$-hop neighbors. 
We report two complementary metrics: \textit{Query Coverage}, defined as $\frac{|T_a \cap T_o|}{|T_a|}$, and \textit{Query Closeness}, defined as $\frac{|T_a \cap T_o|}{|T_o|}$. The coverage measure quantifies the fraction of workload terms that are also mentioned in the ontology; higher values are preferable. The closeness penalizes irrelevant concepts 
not seen in $\Q$; the larger, the better. 

\stitle{Evaluation Method}. Task 2 is triggered by configuration
$(\Q,\O,\RQ,\G,\M,\P)$. If $\RQ=\emptyset$, it invokes
\nltoq to generate $\widehat{\mathcal{RQ}}^{m}$ from $\Q$ and
uses the generated interpretations as the query workload.
It then computes Query Coverage and Query Closeness and may call
\qtoa or \ontoqtoa for answer-accuracy comparison. The computation
pipeline adapts OntoCheck~\cite{ontocheck}, an ontology-assessment
package, to support ontology-grounded SPARQL processing.


\subsection{Task 3: Multimodal Spatiotemporal KGQA}
\label{sec:task3}

Task 3 evaluates the impact of 
multimodal data and queries on answer quality by comparing structured KGQA
workflows (\qtoa or \ontoqtoa) with multimodal counterparts that explicitly
retrieve multimodal information (\multiqtoa or \multiontoqtoa).
Table~\ref{tab:evaluation-metrics} summarizes the answer-quality metrics.

\stitle{Evaluation Metric}. Task 3 uses the answer-quality metrics listed in
Task 1, with a few KGQA-specific additions. Sel. denotes the selected
interpretation-pair score used when multiple generated and reference
interpretations are available. Spatial correctness $S_i$ and temporal
correctness $T_i$ check whether the returned answers preserve the expected
spatial and temporal constraints, and the Spatiotemporal Relevance Score (SRS)
summarizes this spatiotemporal agreement at the answer level.

\stitle{Evaluation Method}. Task 3 is triggered by a configuration
$(\Q,\O,\RQ,\G,\M,\P)$. \geooutagebench executes each reference or generated
interpretation over $\G$ with \qtoa, \ontoqtoa, \multiqtoa, or
\multiontoqtoa, compares $\widehat{\A}_{i}^{\,m}$ with $\A_i$, and reports
macro-averages over the best selected answer pair for each $q_i \in \Q$.

\vspace{-2ex}

\begin{table*}[ht!]
\centering
\caption{Task 1 macro results for NL2SPARQL over 48 GeoOutageBench questions.}
\label{tab:task1-results}
\begin{tabular}{lrrrrrrrrrr}
\toprule
Model & Syntax & Compat. & Exact & Align. & Class F1 & Prop. F1 & Schema F1 & Spatial F1 & Temporal F1 & ST F1 \\
\midrule
GPT-5.5 & 1.000 & 1.000 & 0.000 & 0.745 & 0.716 & 0.679 & 0.678 & 0.870 & 0.715 & 0.662 \\
Gemini 3.1 Pro & 1.000 & 0.979 & 0.021 & 0.749 & 0.834 & 0.703 & 0.737 & 0.783 & 0.675 & 0.578 \\
Claude Opus 4.7 & 1.000 & 1.000 & 0.000 & 0.805 & 0.807 & 0.774 & 0.772 & 0.886 & 0.754 & 0.711 \\
\bottomrule
\end{tabular}
\end{table*}

\begin{table}[t]
\centering
\caption{Assessment of LLMs in Task 1, with finer-grained analysis over spatiotemporal axis. S abbreviates Spatial, T abbreviates Temporal, and ST abbreviates Spatiotemporal.}
\label{tab:task1-axis-results}
\begin{tabular}{llrrrr}
\toprule
Model & Axis & Align. & S F1 & T F1 & ST F1 \\
\midrule
GPT-5.5 & Non-ST & 0.585 & 1.000 & 0.250 & 0.250 \\
GPT-5.5 & S & 0.699 & 0.969 & 0.625 & 0.607 \\
GPT-5.5 & T & 0.751 & 1.000 & 0.626 & 0.676 \\
GPT-5.5 & ST & 0.776 & 0.812 & 0.807 & 0.725 \\
Gemini 3.1 Pro & Non-ST & 0.688 & 1.000 & 0.250 & 0.250 \\
Gemini 3.1 Pro & S & 0.697 & 0.875 & 0.500 & 0.375 \\
Gemini 3.1 Pro & T & 0.734 & 0.750 & 0.609 & 0.485 \\
Gemini 3.1 Pro & ST & 0.771 & 0.738 & 0.780 & 0.682 \\
Claude Opus 4.7 & Non-ST & 0.688 & 1.000 & 0.250 & 0.250 \\
Claude Opus 4.7 & S & 0.727 & 0.875 & 0.625 & 0.500 \\
Claude Opus 4.7 & T & 0.785 & 0.750 & 0.639 & 0.485 \\
Claude Opus 4.7 & ST & 0.842 & 0.891 & 0.863 & 0.850 \\
\bottomrule
\end{tabular}
\end{table}

\begin{table*}[ht!]
\centering
\caption{Task 3 results over 48 questions (Sel.: interpretation selection; SRS: Spatiotemporal Relevance Score)}
\label{tab:kgqa-results}
\begin{tabular}{lrrrrrrrrrr}
\toprule
Model & Prec. & Rec. & Answer F1 & Sel. & Hits@1 & Hits@5 & Hits@10 & MRR & nDCG@10 & SRS \\
\midrule
GPT-5.5 & 0.529 & 0.582 & 0.522 & 0.754 & 0.542 & 0.583 & 0.583 & 0.553 & 0.530 & 0.785 \\
Gemini 3.1 Pro & 0.323 & 0.460 & 0.314 & 0.669 & 0.417 & 0.438 & 0.438 & 0.424 & 0.387 & 0.739 \\
Claude Opus 4.7 & 0.546 & 0.545 & 0.540 & 0.792 & 0.521 & 0.542 & 0.542 & 0.531 & 0.530 & 0.850 \\
\bottomrule
\end{tabular}
\end{table*}

\section{Experimental Evaluation}
\label{sec:eval-assessment}


We next make a case to demonstrate \geooutagebench in assessing 
spatiotemporal KGQA for power outage event analysis. 
We initialize the configuration $\C$ 
by default, where (1) $\O$ and $\G$ refer to  
GeoOutageOnto and GeoOutageKG, respectively, (2) $\Q$ is a
sampled competency query workload spanning all the  
categories, (3) $\M$ includes three frontier LLMs: 
GPT-5.5 with Extended Thinking, 
Gemini 3.1 Pro with Extended Thinking, and 
Claude Opus 4.7 with Extra Effort and Adaptive Thinking; 
and (4) $\P$ is selected to match the 
benchmark tasks. 

\stitle{Environment}.
\geooutagebench is developed in Python with the RDFLib \cite{rdflib} library. 
All tests are conducted on Intel(R) Core(TM) i5-14400F CPU @ 2.50GHz, 32 GB Memory, 10 cores, and 1 8GB NVIDIA(R) RTX(TM) 4060 GPU. Our source code and datasets are made available\footnote{https://github.com/UCF-SAGE/GeoOutageBench}. A live demo of GeoOutageBench, as well as additional resources of the GeoResilience project, will be available on our sandbox website\footnote{https://purl.archive.org/georesilience}.

\stitle{Evaluation-1: Assessing LLM for Power Outage KGQA}. 
The first evaluation measures whether a model $m \in \M$ can translate 
an ambiguous NL question $q_i$ about power-outage and storm-event analysis into ontology-grounded SPARQL. 
We evaluate $\Q_{\mathrm{eval}} \subseteq \Q$ with $|\Q_{\mathrm{eval}}|=48$, where each $q_i$ is associated with 
at most 3 valid SPARQL interpretations in $\RQ_i$. 
\eat{
This one-to-many structure is central to the evaluation: a question may admit multiple valid readings depending on how terms such as ``during,'' ``near,'' ``affected,'' ``high vulnerability,'' or ``outage severity'' are grounded in $\O$ and $\G$. For this reason, exact string match is reported, but it is not the primary indicator of success. More informative scores include interpretation alignment, as well as class, property, schema-linking, and spatiotemporal-constraint correctness. The evaluation employed three foundation model configurations: GPT-5.5 with Extended Thinking, Gemini 3.1 Pro with Extended Thinking, and Claude Opus 4.7 with Extra Effort and Adaptive Thinking.}
Table~\ref{tab:task1-results} summarizes corpus-level results, and
Table~\ref{tab:task1-axis-results} reports a finer-grained ablation assessment over spatial (S), temporal (T), spatiotemporal (ST), and non-spatiotemporal (Non-ST).

\eetitle{Interpretation ambiguity}. 
The 48-question evaluation set contains 72 reference SPARQL interpretations, with each $q_i$ associated with at most three valid readings. Gemini 3.1 Pro generated the fewest interpretations (51), while GPT-5.5 generated 60 and Claude Opus 4.7 generated 61. We treat these counts as an ambiguity diagnostic: fewer generated interpretations reduce downstream execution cost, whereas larger interpretation sets can cover more alternative readings of the same NL question.

\eetitle{Query accuracy}. 
We next report the query accuracy 
(the first three columns of Table~\ref{tab:task1-results}).
All three LLMs in $\M$ generate syntactically valid SPARQL interpretations.
GPT-5.5 and Claude Opus 4.7 also achieve full endpoint compatibility with
$\G$, while Gemini 3.1 Pro achieves 0.979 compatibility. Exact query-text
match is near zero, which is expected because valid interpretations may differ
in variable names, projections, join order, filters, or the selected reading of
an ambiguous NL question.

\eetitle{Alignment \& schema performance.}
Interpretation alignment, defined in Section~\ref{sec:task1}, is therefore a
more informative Task 1 signal than exact match. Claude Opus 4.7 obtains the
highest alignment score (0.805), followed by Gemini 3.1 Pro (0.749) and
GPT-5.5 (0.745). At the schema level, Gemini achieves the highest class F1
(0.834), but Claude leads on property F1 (0.774) and schema-linking F1
(0.772), showing stronger joint recovery of classes and properties.

\eetitle{Spatiotemporal performance.}
Claude Opus 4.7 also leads the corpus-level spatial-, temporal-, and
spatiotemporal-constraint F1 scores, with 0.886, 0.754, and 0.711,
respectively. GPT-5.5 is close on spatial grounding (0.870) and reaches 0.662
ST F1, while Gemini 3.1 Pro is lower on the same spatiotemporal measures
(0.783 spatial F1, 0.675 temporal F1, and 0.578 ST F1). The results
in Table~\ref{tab:task1-axis-results} reinforce this pattern: Claude performs
best on explicitly spatiotemporal questions, with 0.842 alignment and 0.850
ST F1 on the ST axis.

\eetitle{Overall performance}. 
Overall, Claude Opus 4.7 provides the strongest Task 1 performance for NL
spatiotemporal understanding because it combines the best alignment,
schema-linking, and spatiotemporal-constraint scores. GPT-5.5 is the most
competitive alternative on execution reliability, matching Claude on syntax and
endpoint compatibility and remaining close on spatial F1. Gemini 3.1 Pro is
strongest at class recovery and is the only model with a nonzero exact-match
score, but its lower compatibility and spatiotemporal-constraint scores reduce
its overall performance on ontology-grounded spatiotemporal interpretation.
We report details (\eg prompts) 
in Appendix~\ref{app:nl2sparql-prompt}.  

\stitle{Evaluation-2: Ontology Utility Evaluation}. 
We evaluated the coverage of $\O$ over the SPARQL workload using OntoCheck's task-based framework \cite{ontocheck}. The query workload contains 72 query interpretations. 
For the case of GeoOutageOnto $\O$, 
the Query Coverage is 1.0: every extracted \texttt{goo:} 
domain term in the query workload has been declared in $\O$, leaving no extracted GeoOutageOnto term unanchored in the schema. This result is expected given the co-design of GeoOutageOnto and the benchmark workload. The Query Closeness, which measures schema-utilization, ranges from 0.0196 for simple single-predicate lookups to 0.1373 for the most compositionally complex queries, with a mean of 0.0673. This range reflects benchmark heterogeneity rather than an ontology weakness: simple queries activate a small subset of the vocabulary, whereas multi-hop, spatiotemporal, and aggregative queries require broader schema coverage.

\eetitle{Case study on competency queries}. 
We present three query-case analyses of LLM-generated interpretations and compare their answers. 
(1) Queries 70--72 (Average Query Closeness = 0.1078) 
are different interpretations for the 
same NL question ``\textit{Which counties in the hurricane-prone state had high socioeconomic vulnerability in a recent year, and a dozen or so wind-related storm events recorded that year?}'' with different predicates posed on state, year, vulnerability threshold, event type, and count; 
all conform to the same GeoOutageOnto vocabulary. 
(2) Queries 22--24 (Average Query Closeness = 0.0490) resolve ``\textit{around Hurricane Ian}'' as a 15-day impact window (2022-09-23 to 2022-10-08), a two-month disruption period (2022-08-28 to 2022-10-28), or the Florida landfall date (2022-09-28 to 2022-09-29 UTC), while consistently grounding Lee County severity maps through ontology-defined imagery classes and spatial, temporal, and map-association properties. (3) Queries 37--39 (Average Query Closeness = 0.0686) resolve ``\textit{within the general area of Cleveland County}'' to within a 10 km radius, a 50 km radius, and a buffered Moore tornado event window, respectively.

\vspace{.5ex}
These observations support two conclusions. First, the consistently high query coverage 
indicates that $\O$ covers all extracted \texttt{goo:} domain terms in the query workload, 
suggesting that GeoOutageOnto provides a grounded schema reference for power-outage 
analysis over the spatiotemporal GeoOutageKG $\G$. Second, query-level closeness variation 
shows that different workload types activate different portions of $\O$: simple lookups use 
small schema subsets, while multi-hop, spatiotemporal, multimodal, and aggregative queries 
use broader vocabulary. The per-query scores make this variation visible, including the 
multimodal evidence interpretations in queries 46--48 and the alternative interpretations 
of the ambiguous outage-and-storm question in queries 70--72. A full list of per-query 
scores is reported in Table~\ref{tab:ontocheck}, Appendix~\ref{app:evaluation-metrics}.

\stitle{Evaluation-3: Multimodal KGQA}. 
The third evaluation executes $\widehat{\RQ}_{i}^{\,m}$ against $\G$ and compares the generated answer set $\widehat{\A}_{i}^{\,m}$ against ground-truth answers $\A_i$. This task differs from query accuracy in Task 1, because a query can be syntactically valid and structurally plausible yet still return an irrelevant answer. 

\vspace{.5ex}
Table~\ref{tab:kgqa-results} emphasizes that answer-level evaluation over $\A_i$ is more discriminative than syntax or executability. Claude Opus 4.7 obtains the highest answer F1 and SRS, while GPT is comparable in answer F1 and slightly higher on Hits@1; Gemini 3.1 Pro lags in answer F1 despite strong Task 1 syntax and class-recovery scores. This gap shows that correct ontology vocabulary alone does not ensure KGQA correctness: small errors in property choice, temporal filtering, aggregation, or administrative joins can preserve a plausible query form while changing the returned answers.

\vspace{.5ex}
The relatively higher SRS scores for GPT and Claude indicate that many generated queries preserve the broad spatial and temporal intent even when the answer rows differ from gold results. This is typically because the query relaxes predicates, omits an aggregation, or returns less relevant evidence such as outage records rather than derived outage maps. Complexity and template breakdowns show the same observation. Two-hop questions are much easier than multi-hop questions for GPT (0.992 vs. 0.310 answer F1) and Claude (0.833 vs. 0.359), while Gemini drops most sharply on multi-hop questions (0.151). Template-level results show that direct lookup and outage-ranking queries are generally more reliable, while storm-event/SVI threshold queries are model-dependent, with strong performance for Claude but weaker results for Gemini. In contrast, hurricane-landfall ranking and event-sequence questions remain difficult because they require coordinated event ranking, spatial aggregation, storm semantics, and a proper 
setting of spatiotemporal-sensitive vulnerability thresholds.

\begin{figure}[t]
\centering
\includegraphics[width=0.95\columnwidth,totalheight=0.82\textheight,keepaspectratio]{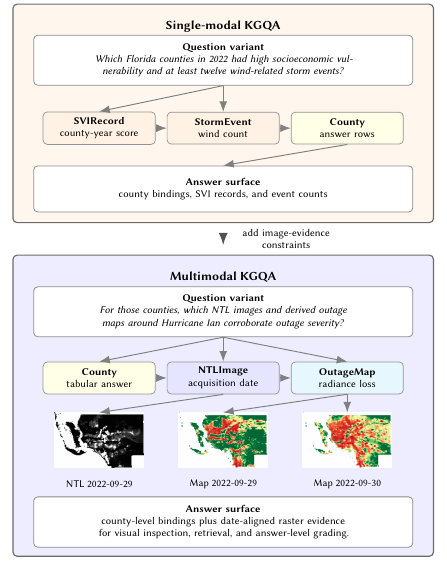}
\Description{Two-panel case-study figure comparing single-modal and multimodal KGQA. The single-modal panel asks for Florida counties in 2022 with high socioeconomic vulnerability and at least twelve wind-related storm events, returning SVI records, storm-event counts, and county rows. The multimodal panel requests nighttime-light images and derived outage maps around Hurricane Ian for those counties, returning county bindings along with image assets from September 29 and September 30, 2022.}
\caption{Single- versus multimodal querying for the outage-and-storm case study. The multimodal variant supplements county-level tabular answers with date-aligned NTL images and outage maps, enabling evidence-aware visual inspection and richer answer-level assessment.}
\label{fig:single-vs-multimodal-case-study}
\end{figure}

\eetitle{Single Modal vs. Multimodal Querying: 
A Case Study}. 
\eat{
\comwu{We have room to 
add a visual illustration to show benefit of 
multimodal KG: give a multimodal NL query, with 
answers in the KG that have content 
extracted from satellite image, and 
data frames, and compare it with a 
single modal counterpart that only returns 
entities curated from \eg population or environment documents. Show the difference and 
discuss further applications.}
}
We use the ambiguous outage-and-storm question from
Example~\ref{exa-motivation} to illustrate what changes when KGQA moves from a
single-modal, tabular answer surface to a multimodal one. A single-modal
variant asks:
\textit{``Which Florida counties in 2022 had high socioeconomic vulnerability
and at least twelve wind-related storm events?''}
This query returns county, CDC SVI, and NOAA Storm Events bindings, but no remote-sensing evidence of the outage footprint, leading to potentially biased or less interpretable results.

\vspace{.5ex}
The multimodal variant keeps the same county-level analytic intent but adds an
evidence requirement:
\textit{``For those counties, which nighttime-light images and derived outage
maps around Hurricane Ian corroborate outage severity?''}
In this setting, \multiqtoa or \multiontoqtoa returns not only the county, SVI,
and storm-event bindings, but also date-aligned \texttt{goo:NTLImage} and
\texttt{goo:OutageMap} assets linked through the same county and time
constraints. Figure~\ref{fig:single-vs-multimodal-case-study} displays how additional image evidence changes the assessment target: the
benchmark can grade entity-level correctness while also checking whether the
retrieved multimodal bundle supports the intended spatiotemporal explanation.
This pattern is useful for downstream resilience applications in which analysts
need both ranked county answers and inspectable evidence for outage extent,
recovery timing, or vulnerability-aware resource prioritization.




\section{Conclusions}
\label{sec:conclusions}

We have introduced \geooutagebench, a domain-specific KGQA benchmark for evaluating LLM-based spatiotemporal KGQA for outage, social-vulnerability, and storm-resilience analysis. The benchmark curates multimodal data 
into a geospatiotemporal knowledge graph 
(GeoOutageKG) and ontology (GeoOutageOnto). 
It supports automatic generation of 
diverse competency queries 
that cover seven geospatiotemporal templates 
and provides a built-in library of primitive 
operators for efficient assessment. 
It provides comprehensive performance metrics to assess 
LLMs' spatiotemporal understanding, 
usability and utility of ontologies, 
and multimodal KGQA. Our experimental evaluation 
verifies that Claude Opus 4.7 achieves the best 
overall performance in NL spatiotemporal 
understanding, while GPT-5.5 remains 
competitive on executability and 
spatial grounding, and Gemini 3.1 Pro 
is strongest on class recovery. Meanwhile, 
our query-level and finer-grained assessment  
verifies the effectiveness of 
GeoOutageOnto and multimodal 
GeoOutageKG for improving 
the answer accuracy for power outage 
analysis. 

Future work will broaden the assessment to richer query semantics, including cross-modal and cross-model (hybrid raster-and-tabular) retrieval, counterfactual spatiotemporal reasoning, and other causes of energy disruption such as cyberattacks and grid overload. We will also extend the Black Marble and EAGLE-I layers beyond the sampled Florida counties.

\begin{acks}
This material is based upon research in the Materials Data Science for Stockpile Stewardship Center of Excellence (MDS3-COE), and supported by the Department of Energy's National Nuclear Security Administration under Award Number(s) DE-NA0004104.
The authors thank the CWRU University Technology center and the UCF Advanced Research Computing Center for their High Performance Computing (HPC) resources, which were utilized in this work.
\end{acks}

\section*{Declaration of the use of Generative AI}
The usage of Large Language Models and Agents, including ChatGPT, Google Gemini, Claude, and coding agents such as OpenAI Codex, was employed to automate code and experiment script generation, with human refinement and editing. LLMs were also used to help in refining portions of text, tables, and figures.

{%
\RestoreLaTeXAccents
\bibliographystyle{ACM-Reference-Format}
\bibliography{refs}
}

\appendix

\section{Appendix}
\label{app:appendix}

\subsection{Query-Interpretation: Prompts}
\label{app:nl2sparql-prompt}

Task 1 uses ontology-grounded prompts to generate the query interpretations evaluated in Sections~\ref{sec:benchmark_tasks} and~\ref{sec:eval-assessment}. This process implements the operator 
\nltoq with the following steps. 
(1) The LLM receives a JSON prediction template with benchmark metadata, the GeoOutageOnto ontology in Turtle, and a compact vocabulary file containing core classes, properties, prefixes, and domain/range; gold SPARQL queries are not provided. (2) The interpreter shares the template and consults the LLM to fill \texttt{sparql\_query} with the best default interpretation. To mitigate ambiguity, it includes the query set (a memory space \texttt{sparql\_interpretations}) containing syntactically correct SPARQL queries.

The prompt constrains the generated SPARQL interpretation to align with the ontology vocabulary when available, requires distinct interpretations for different spatial, temporal, semantic, or spatiotemporal readings, and discourages variants that merely rename variables or reorder triples. It also standardizes SPARQL style through required prefixes, bound projected variables, ISO 8601 \texttt{xsd:dateTime} filters, half-open day-level ranges, aggregation and ranking rules, and namespace requirements. 

Figure~\ref{fig:appendix-nl2sparql-prompt} summarizes the operative structure of the prompt; the full prompt is available in the GitHub repository. Internally, the prompt asks the model to identify answer type, ontology class, connecting properties, spatial or temporal filters, aggregation, ranking, multimodal evidence, and ambiguity, but the returned output is only the completed JSON prediction file.

\subsection{Evaluation Metrics}
\label{app:evaluation-metrics}

\begin{figure}[t]
\centering
\begin{minipage}{0.99\linewidth}
\begin{lstlisting}[basicstyle=\ttfamily\scriptsize,breaklines=true,frame=single]
Role: expert GeoSPARQL, temporal SPARQL, and ontology-grounded query generation system.

Inputs:
1. GeoOutageBench prediction JSON with empty sparql_query fields.
2. Full ontology in Turtle.
3. Allowed classes, properties, prefixes, and domain/range hints.

Core rules:
- Use only the provided ontology/schema context.
- Do not use gold/reference SPARQL.
- Preserve existing JSON fields.
- Fill sparql_query with the default interpretation.
- Add sparql_interpretations only for answer-changing ambiguity.
- Return only valid JSON.
- Generate executable SPARQL using allowed prefixes.

Ambiguity rules:
- Generate separate queries for materially different spatial, temporal, semantic, or spatiotemporal readings.
- Do not create variants that only rename variables or reorder triples.
- Keep interpretations small, usually 1--3.

Output for each interpretation:
label, ambiguity_dimensions, rationale, sparql_query.
\end{lstlisting}
\end{minipage}
\Description{Condensed structure of the NL2SPARQL prompt used to produce GeoOutageBench query interpretations.}
\caption{Condensed structure of the \nltoq prompt used to produce GeoOutageBench query interpretations.}
\label{fig:appendix-nl2sparql-prompt}
\end{figure}

GeoOutageBench evaluates formal query generation, ontology utility, and answer-level spatiotemporal KGQA. We use $\widehat{\RQ}_{i}^{\,m}$ for the realized generated interpretation set of model $m$ on question $q_i$ and $\RQ_i$ for the reference interpretation set. Unless otherwise stated, corpus scores are macro-averages over the $N$ evaluated questions:

\begin{equation}
\overline{\mu}=\frac{1}{N}\sum_{i=1}^{N}\mu_i .
\label{eq:macro-average}
\end{equation}

Several metrics use the same set-overlap precision, recall, and F1 form. For a generated set $\widehat{X}_i$, a gold set $X_i$, and a matched-correct count $c_i$, we compute

\begin{equation}
\begin{aligned}
P_i&=\frac{c_i}{|\widehat{X}_i|}, &
R_i&=\frac{c_i}{|X_i|},\\
F1_i&=
\begin{cases}
\dfrac{2P_iR_i}{P_i+R_i}, & \text{if } P_i+R_i\neq 0,\\
0, & \text{if } P_i+R_i=0.
\end{cases}
\end{aligned}
\label{eq:prf}
\end{equation}

When both generated and gold sets are empty, the corresponding feature-family score is defined as $1$; when the prediction is empty and the gold set is non-empty, it is $0$.

\stitle{Task 1 metrics}. The Task 1 evaluator first normalizes each prediction into a deduplicated list of query records, accepting either a single \texttt{sparql\_query} or richer ambiguity-aware fields such as \texttt{sparql\_interpretations}. Each generated interpretation is scored against each reference interpretation, and the best aligned pair gives the question-level structural score. Class F1 is based on qnames used as \texttt{rdf:type} targets, property F1 uses graph-pattern predicates excluding \texttt{rdf:type}, spatial F1 uses GeoSPARQL/WGS84, geometry, coordinate, distance, and spatial-predicate markers, and temporal F1 uses date/dateTime literals and temporal filter markers. Axis-level results group questions by non-spatiotemporal, spatial, temporal, or spatiotemporal annotations.

Let $\nu(r)$ denote normalized SPARQL text after removing prefixes and comments, collapsing whitespace, and lowercasing. Let $\widehat{N}_i^m=\{\nu(\hat{r})\mid \hat{r}\in\widehat{\RQ}_{i}^{\,m}\}$ and $N_i=\{\nu(r)\mid r\in\RQ_i\}$. Syntax validity is the fraction of generated interpretations that pass the coarse SELECT/ASK and WHERE/braces check:

\begin{equation}
\mathrm{Syn}_i^m=
\frac{1}{|\widehat{\RQ}_{i}^{\,m}|}
\sum_{\hat{r}\in\widehat{\RQ}_{i}^{\,m}}
\mathbf{1}\{\hat{r}\text{ is syntactically valid}\}.
\label{eq:syntax}
\end{equation}
If $\widehat{\RQ}_{i}^{\,m}=\emptyset$, syntax validity is defined as $0$.

Compatibility (also referred to as executability) receives credit when at least one generated interpretation executes successfully over the configured graph or endpoint:

\begin{equation}
\begin{aligned}
\mathrm{Compat}_i^m
=\mathbf{1}\{&\exists \hat{r}\in\widehat{\RQ}_{i}^{\,m}:\\
&\hat{r}\text{ executes over }\G\}.
\end{aligned}
\label{eq:exec}
\end{equation}

Exact original-query match receives credit when the original source query $r_{i0}$ appears anywhere in the generated interpretation set, while ambiguity-aware matching compares the full interpretation sets:

\begin{equation}
\begin{aligned}
\mathrm{Exact}_i^m
&=\mathbf{1}\{\nu(r_{i0})\in\widehat{N}_i^m\},\\
\mathrm{Any}_i^m
&=\mathbf{1}\{\widehat{N}_i^m\cap N_i\neq\emptyset\}.
\end{aligned}
\label{eq:exact-any}
\end{equation}

Interpretation recall is

\begin{equation}
\mathrm{IRec}_i^m=
\frac{|\widehat{N}_i^m\cap N_i|}{|N_i|}.
\label{eq:interpretation-recall}
\end{equation}

For structural alignment, Eq.~\ref{eq:prf} is applied to extracted feature sets: classes, properties, spatial constraints, and temporal constraints. The selected interpretation pair maximizes the average of these four F1 scores:

\begin{equation}
\begin{aligned}
\mathrm{Align}_i^m=
&\max_{\substack{\hat{r}\in\widehat{\RQ}_{i}^{\,m}\\ r\in\RQ_i}}
\frac{1}{4}\bigl(
\mathrm{ClassF1}_{\hat{r},r}
+\mathrm{PropF1}_{\hat{r},r}\\
&\hspace{7.5em}
+\mathrm{SF1}_{\hat{r},r}
+\mathrm{TF1}_{\hat{r},r}
\bigr).
\end{aligned}
\label{eq:alignment}
\end{equation}

Schema-linking F1 and spatiotemporal-constraint F1 use harmonic means:

\begin{equation}
\begin{aligned}
\mathrm{SchemaF1}_i^m=
\begin{cases}
\dfrac{2\cdot \mathrm{ClassF1}_i^m\cdot \mathrm{PropF1}_i^m}
{\mathrm{ClassF1}_i^m+\mathrm{PropF1}_i^m},
& \begin{gathered}
\text{if } \mathrm{ClassF1}_i^m\\
+\mathrm{PropF1}_i^m\neq 0,
\end{gathered}\\
0, & \text{otherwise,}
\end{cases}
\end{aligned}
\label{eq:schema-f1}
\end{equation}

\begin{equation}
\begin{aligned}
\mathrm{STF1}_i^m = 
\begin{cases}
\dfrac{2 \cdot SF1_i^m \cdot TF1_i^m}{SF1_i^m + TF1_i^m},
& \text{if } SF1_i^m + TF1_i^m \neq 0, \\
0, & \text{if } SF1_i^m + TF1_i^m = 0.
\end{cases}
\end{aligned}
\label{eq:stf1}
\end{equation}

\begin{table}[t]
\centering
\caption{Task-Based Assessment of GeoOutageOnto through OntoCheck \cite{ontocheck}.}
\label{tab:ontocheck}
\small
\setlength{\tabcolsep}{2pt}
\renewcommand{\arraystretch}{0.92}
\ACMmaxwidth{\columnwidth}{%
\begin{tabular}{@{}rc@{\hspace{0.65em}}rc@{\hspace{0.65em}}rc@{\hspace{0.65em}}rc@{\hspace{0.65em}}rc@{\hspace{0.65em}}rc@{}}
\toprule
Q & Close. & Q & Close. & Q & Close. & Q & Close. & Q & Close. & Q & Close. \\
\midrule
1  & 0.0294 & 13 & 0.0686 & 25 & 0.0392 & 37 & 0.0686 & 49 & 0.0784 & 61 & 0.0882 \\
2  & 0.0392 & 14 & 0.0196 & 26 & 0.0392 & 38 & 0.0686 & 50 & 0.0784 & 62 & 0.1373 \\
3  & 0.0294 & 15 & 0.0196 & 27 & 0.0392 & 39 & 0.0686 & 51 & 0.0784 & 63 & 0.1275 \\
4  & 0.0294 & 16 & 0.0196 & 28 & 0.0294 & 40 & 0.0392 & 52 & 0.0490 & 64 & 0.0686 \\
5  & 0.0392 & 17 & 0.0392 & 29 & 0.0686 & 41 & 0.0686 & 53 & 0.0490 & 65 & 0.0686 \\
6  & 0.0392 & 18 & 0.0392 & 30 & 0.0294 & 42 & 0.0882 & 54 & 0.1176 & 66 & 0.0980 \\
7  & 0.0196 & 19 & 0.0588 & 31 & 0.0490 & 43 & 0.0882 & 55 & 0.0686 & 67 & 0.0980 \\
8  & 0.0196 & 20 & 0.0588 & 32 & 0.0588 & 44 & 0.0882 & 56 & 0.1176 & 68 & 0.0980 \\
9  & 0.0196 & 21 & 0.0588 & 33 & 0.0490 & 45 & 0.0882 & 57 & 0.0686 & 69 & 0.0980 \\
10 & 0.0196 & 22 & 0.0490 & 34 & 0.1078 & 46 & 0.1078 & 58 & 0.1078 & 70 & 0.1078 \\
11 & 0.0196 & 23 & 0.0490 & 35 & 0.1078 & 47 & 0.1078 & 59 & 0.0980 & 71 & 0.1078 \\
12 & 0.0490 & 24 & 0.0490 & 36 & 0.1078 & 48 & 0.1078 & 60 & 0.1275 & 72 & 0.1078 \\
\midrule
\multicolumn{12}{c}{\textbf{Average Closeness}: \textbf{0.0673}} \\
\bottomrule
\end{tabular}%
}%
\end{table}

\stitle{Task 2 metrics}. OntoCheck \cite{ontocheck} constructs a task term set $T_a(r_{ij})$ from extracted \texttt{goo:} terms in each SPARQL interpretation $r_{ij}\in\RQ_i$ and compares it with the scoped ontology term set $T_o$. Query Coverage (or Relevance) measures how much workload vocabulary is declared by the ontology; Query Closeness (or Accuracy) measures how densely the scoped ontology vocabulary is activated by the query:

\begin{equation}
\begin{aligned}
\mathrm{Cov}_{ij}
&=\frac{|T_a(r_{ij})\cap T_o|}{|T_a(r_{ij})|},\\
\mathrm{Clo}^{onto}_{ij}
&=\frac{|T_a(r_{ij})\cap T_o|}{|T_o|}.
\end{aligned}
\label{eq:ontocheck}
\end{equation}

The full list of Closeness scores for all 72 queries in the ground-truth set is listed in Table~\ref{tab:ontocheck} (all queries returned a Coverage of 1.00, hence are not shown).

\stitle{Task 3 metrics}. Task 3 treats Task 1 outputs as query workload to be processed. Each SPARQL interpretation is executed against GeoOutageKG, using GraphDB~\cite{GraphDB} for the reported experiments, and gold answers are obtained by executing 
ground-truth interpretations, or by using stored human-validated benchmark answers if all interpretations return empty results. Returned bindings are normalized by compacting IRIs, simplifying datatype and language-tag variants, normalizing numeric and UTC date/time values, and aligning common aliases such as \texttt{ntl}/\texttt{image}, \texttt{rec}/\texttt{record}, and \texttt{num}/\texttt{numOutages}.

Answer scoring uses Eq.~\ref{eq:prf} with $\widehat{X}_i=\widehat{\A}_{i}^{\,m}$, $X_i=\A_i$, and $c_i$ equal to the number of generated rows matched to distinct gold rows after normalization. For ranked outputs, let $y_{i\ell}\in\{0,1\}$ indicate whether the row at rank $\ell$ is relevant. We compute

\begin{equation}
\begin{aligned}
\mathrm{Hits@}k_i
&=\mathbf{1}\Bigl\{\sum_{\ell=1}^{k}y_{i\ell}>0\Bigr\},\\
\mathrm{RR}_i=
\begin{cases}
\dfrac{1}{\min\{\ell:y_{i\ell}=1\}}, & \text{if any } y_{i\ell}=1,\\
0, & \text{otherwise,}
\end{cases}
\end{aligned}
\label{eq:hits-rr}
\end{equation}

and report MRR as the macro-average of $\mathrm{RR}_i$. Ranking quality at depth 10 is

\begin{equation}
\begin{aligned}
\mathrm{DCG@10}_i
&=\sum_{\ell=1}^{10}\frac{y_{i\ell}}{\log_2(\ell+1)},\\
\mathrm{nDCG@10}_i
&=\frac{\mathrm{DCG@10}_i}{\mathrm{IDCG@10}_i}
\quad \text{if } \mathrm{IDCG@10}_i>0,\\
&=0 \quad \text{otherwise.}
\end{aligned}
\label{eq:ndcg}
\end{equation}

Spatial correctness $S_i$ and temporal correctness $T_i$ are defined from explicit annotations when available; otherwise, they are derived from overlap between the selected generated and gold SPARQL constraints. Dimensions not required by the question receive full credit, while required dimensions fall back to answer F1 only when no query-level signal is available. The Spatiotemporal Relevance Score is

\begin{equation}
\begin{aligned}
\mathrm{SRS}_i =
\begin{cases}
\dfrac{2 S_i T_i}{S_i + T_i}, & \text{if } S_i + T_i \neq 0, \\
0, & \text{if } S_i + T_i = 0.
\end{cases}
\end{aligned}
\label{eq:srs}
\end{equation}

With multiple generated and reference interpretations, answer-level alignment selects the generated--gold pair maximizing the interpretation-selection score

\begin{equation}
\begin{aligned}
\mathrm{Sel}_i^m=
\frac{1}{4}\bigl(
\mathrm{AnswerF1}_i^m+S_i+T_i\\
\hspace{5.2em}+\mathrm{SRS}_i
\bigr).
\end{aligned}
\label{eq:selection-score}
\end{equation}

\eat{
\subsection{Full OntoCheck Scores}
\label{app:task2-methodology}

Table~\ref{tab:ontocheck} reports the full per-query OntoCheck scores for the 72-query interpretation workload. Coverage is 1.0 for every query, indicating that all extracted \texttt{goo:} workload terms are declared in GeoOutageOnto. Closeness varies with query complexity: simple lookups activate a narrow subset of the ontology, whereas multi-hop spatiotemporal, multimodal, and scenario-oriented queries use a broader vocabulary.
}



\end{document}